\documentclass{article}

\usepackage{microtype}
\usepackage{graphicx}
\usepackage{subfigure}
\usepackage{booktabs} 

\usepackage{hyperref}
\usepackage{multirow}

\usepackage[accepted]{icml2019}

\icmltitlerunning{Machine Learning for Clinical Predictive Analytics}

\begin{document}

\twocolumn[
\icmltitle{Machine Learning for Clinical Predictive Analytics}



\icmlsetsymbol{equal}{*}

\begin{icmlauthorlist}
\icmlauthor{Wei-Hung Weng}{goo}
\end{icmlauthorlist}

\icmlaffiliation{goo}{MIT CSAIL, Cambridge, MA, USA}

\icmlcorrespondingauthor{Wei-Hung Weng}{ckbjimmy@mit.edu}

\icmlkeywords{Machine Learning, Clinical}

\vskip 0.3in
]



\printAffiliationsAndNotice{}  

\begin{abstract}
\begin{itemize}
    \item Understand the basics of machine learning techniques and the reasons behind why they are useful for solving clinical prediction problems.
    \item Understand the intuition behind some machine learning models, including regression, decision trees, and support vector machines.
    \item Understand how to apply these models to clinical prediction problems using publicly available datasets via case studies.
\end{itemize}
\end{abstract}

\section{Machine Learning for Healthcare}
\label{s1}

\subsection{Introduction}
In this chapter, we provide a brief overview of applying machine learning techniques for clinical prediction tasks. We begin with a quick introduction to the concepts of machine learning, and outline some of the most common machine learning algorithms. Next, we demonstrate how to apply the algorithms with appropriate toolkits to conduct machine learning experiments for clinical prediction tasks.

This chapter is composed of five sections. First, we will explain why machine learning techniques are helpful for researchers in solving clinical prediction problems (section \ref{s1}). Understanding the motivations behind machine learning approaches in healthcare are essential, since precision and accuracy are often critical in healthcare problems, and everything from diagnostic decisions to predictive clinical analytics could dramatically benefit from data-based processes with improved efficiency and reliability. In the second section, we will introduce several important concepts in machine learning in a colloquial manner, such as learning scenarios, objective/target function, error and loss function and metrics, optimization and model validation, and finally a summary of model selection methods (section \ref{s2}). These topics will help us utilize machine learning algorithms in an appropriate way. Following that, we will introduce some popular machine learning algorithms for prediction problems (section \ref{s3}). For example, logistic regression, decision tree and support vector machine. Then, we will discuss some limitations and pitfalls of using the machine learning approach (section \ref{s4}). Lastly, we will provide case studies using real intensive care unit (ICU) data from a publicly available dataset, PhysioNet Challenge 2012, as well as the breast tumor data from Breast Cancer Wisconsin (Diagnostic) Database, and summarize what we have mentioned in the chapter (section \ref{s5}).

\subsection{Why machine learning?}
Machine learning is an interdisciplinary field which consists of computer science, mathematics, and statistics. It is also an approach toward building intelligent machines for artificial intelligence (AI). Different from rule-based symbolic AI, the idea of utilizing machine learning for AI is to learn from data (examples and experiences). Instead of explicitly programming hand-crafted rules, we construct a model for prediction by feeding data into a machine learning algorithm, and the algorithm will learn an optimized function based on the data and the specific task. Such data-driven methodology is now the state-of-the-art approach of various research domains, such as computer vision~\cite{krizhevsky2012imagenet}, natural language processing (NLP)~\cite{yala2017using}, and speech to text translation~\cite{wu2016google,chung2018unsupervised,chung2019towards}, for many complex real-world applications. 

Due to the increased popularity of the electronic health record (EHR) system in recent years, massive quantities of healthcare data have been generated~\cite{henry2016adoption}. Machine learning for healthcare therefore becomes an emerging applied domain. Recently, researchers and clinicians have started applying machine learning algorithms to solve the problems of clinical outcome prediction~\cite{ghassemi2014unfolding}, diagnosis~\cite{gulshan2016development,esteva2017dermatologist,liu2017detecting,chung2017learning,nagpal2018development}, treatment and optimal decision making~\cite{raghu2017continuous,weng2017representation,komorowski2018artificial} using data in different modalities, such as structured lab measurements~\cite{pivovarov2015learning}, claims data~\cite{doshi2014comorbidity,pivovarov2015learning,choi2016learning}, free texts~\cite{pivovarov2015learning,weng2018mapping,weng2019unsupervised}, images~\cite{gulshan2016development,esteva2017dermatologist,bejnordi2017deep,chen2019augmented}, physiological signals~\cite{lehman2018representation}, and even cross-modal information~\cite{hsu2018unsupervised,liu2019clinically}.

Instead of traditional ad-hoc healthcare data analytics, which usually requires expert-intensive efforts for collecting data and designing limited hand-crafted features, machine learning-based approaches help us recognize patterns inside the data and allow us to perform personalized clinical prediction with more generalizable prediction models~\cite{gehrmann2018comparing}. They help us maximize the utilization of massive but complex EHR data. In this chapter, we will focus on how to tackle clinical prediction problems using a machine learning-based approach.

\section{General Concepts of Learning}
\label{s2}

\subsection{Learning scenario for clinical prediction}
We start with how to frame your clinical problem into a machine learning prediction problem with a simple example. Assuming that you want to build a model for predicting the mortality of ICU patients with continuous renal replacement therapy and you have a large ICU database, which includes hundreds of variables such as vital signs, lab data, demographics, medications, and even clinical notes and reports, the clinical problem can be reframed as a task: ``Given data with hundreds of input variables, I want to learn a model from the data that can correctly make a prediction given a new datapoint.'' That is, the output of the function (model) should be as close as possible to the outcome of what exactly happened (the ground truth). Machine learning algorithm is here to help you to find the best function from a set of functions. This is a typical machine learning scenario, which is termed supervised learning. In such a case, you may do the following steps:
\begin{itemize}
    \item Define the outcome of your task
    \item Consult with domain experts to identify important features/variables
    \item Select an appropriate algorithm (or design a new machine learning algorithm) with a suitable parameter selection
    \item Find an optimized model with a subset of data (training data) with the algorithm
    \item Evaluate the model with another subset of data (testing data) with appropriate metrics
    \item Deploy the prediction model on real-world data
\end{itemize}
At the end of the chapter, we will show an exercise notebook that will help you go through the concepts mentioned above.

\subsection{Machine learning scenarios}
There are many machine learning scenarios, such as supervised learning, unsupervised learning, semi-supervised learning, reinforcement learning, and transfer learning. We will only focus on the first two main categories, supervised learning and unsupervised learning. Both of the scenarios expect to learn from the underlying data distribution, or to put it simply, find patterns inside data. The difference between them is that you have annotated data under the supervised scenario but only unlabelled data under unsupervised learning scenario.

\subsubsection{Supervised learning}
Supervised learning is the most common scenario for practical machine learning tasks if the outcome is well-defined, or example, if you are predicting patient mortality, hospital length of stay, or drug response. In general, the supervised learning algorithm will try to learn how to build a classifier for predicting the outcome variable $y$ given input $x$, which is a mapping function $f$ where $y=f(x)$. The classifier will be built by an algorithm along with a set of data $\{x_1, ..., x_n\}$ with the corresponding outcome label $\{y_1, ..., y_n\}$.Supervised learning can be categorized by two criteria, either by type of prediction or by type of model. First, it can be separated into regression or classification problems. For predicting continuous outcomes, using regression methods such as linear regression is suitable. For class prediction, classification algorithms such as logistic regression, naive Bayes, decision trees or support vector machines (SVM)~\cite{cortes1995support} will be a better choice. For example, linear regression is suitable for children height prediction problem whereas SVM is better for binary mortality prediction.

Regarding the goal of the learning process, a discriminative model such as regression, trees and SVMs can learn the decision boundary within the data.However, a generative model like naive Bayes will learn the probability distributions of the data.

\subsubsection{Unsupervised learning}
Without corresponding output variables ($y$), the unsupervised learning algorithms discover latent structures and patterns directly from the given unlabeled data $\{x_1, ..., x_n\}$.

There is no ground truth in the unsupervised learning, therefore, the machine will only find associations or clusters inside the data.For example, we may discover hidden subtypes in a disease using an unsupervised approach~\cite{ghassemi2014unfolding}.

\subsubsection{Other scenario}
Other scenarios such as reinforcement learning (RL) frame a decision making problem into a computer agent interaction with a dynamic environment~\cite{silver2016mastering}, in which the agent attempts to reach the best reward based on feedback when it navigates the state and action space.
Using a clinical scenario as an example, the agent (the RL algorithm) will try to improve the model parameters based on iteratively simulating the state (patient condition) and action (giving fluid or vasopressor for hypotension), obtain the feedback reward (mortality or not), and eventually converge to a model that may yield optimal decisions~\cite{raghu2017continuous}.

\subsection{Find the best function}
To estimate and find the best mapping function in the above scenarios, the process of optimization is needed. However, we do need to define some criteria to tell us how well the function (model) can predict the task. Therefore, we need a loss function and a cost function (objective function) for this purpose.

Loss function defines the difference between the output of model $y$ and the real data value $\hat{y}$. Different machine learning algorithms may use different loss functions, for example, least squared error for linear regression, logistic loss for logistic regression, and hinge loss for SVM (Table~\ref{tab:loss}). Cost function is the summation of loss functions of each training data point. Using loss functions, we can define the cost function to evaluate model performance. Through loss and cost functions, we can compute the performance of functions on the whole dataset.

In unsupervised learning setting, the algorithms have no real data value to compute the loss function. In such case, we can use the input itself as the output and compute the difference between input and output. For example, we use reconstruction loss for autoencoder, a kind of unsupervised learning algorithms, to evaluate whether the model can well reconstruct the input from hidden states inside the model. 

\begin{table*}[htbp]
\centering
\resizebox{\textwidth}{!}{
\begin{tabular}{cccl}
\toprule
Task & Error type & Loss function & \multicolumn{1}{c}{Note} \\
\midrule
\multirow{2}{*}{Regression} & Mean-squared error & $\frac{1}{n} \sum_{i=1}^n (y_i - \hat{y_i})^2$ & \begin{tabular}[c]{@{}l@{}}Easy to learn but sensitive to outliers \\ (MSE, L2 loss)\end{tabular} \\ \cline{2-4} 
 & Mean absolute error & $\frac{1}{n} \sum_{i=1}^n |y_i - \hat{y_i}|$ & \begin{tabular}[c]{@{}l@{}}Robust to outliers but not differentiable \\ (MAE, L1 loss)\end{tabular} \\
 \midrule
\multirow{3}{*}{Classification} & \begin{tabular}[c]{@{}c@{}} Cross entropy \\ = Log loss \end{tabular} & \begin{tabular}[c]{@{}c@{}} $- \frac{1}{n} \sum_{i=1}^n [y_i \log(\hat{y_i}) + (1-y_i) \log(1-\hat{y_i})]$ \\ = $- \frac{1}{n} \sum_{i=1}^n p_i \log q_i$ \end{tabular} & \begin{tabular}[c]{@{}l@{}}Quantify the difference between \\ two probability distributions\end{tabular} \\ \cline{2-4} 
 & Hinge loss & $\frac{1}{n} \sum_{i=1}^n max(0, 1 - y_i \hat{y_i})$ & For support vector machine \\ \cline{2-4} 
 & KL divergence & $D_{KL}(p||q) = \sum_{i} p_i (\log \frac{p_i}{q_i})$ & \begin{tabular}[c]{@{}l@{}}Quantify the difference between \\ two probability distributions\end{tabular} \\
\bottomrule
\end{tabular}
}
\caption{Examples of commonly-used loss functions in machine learning.}
\label{tab:loss}
\end{table*}

There is a mathematical proof for this learning problem to explain why machine learning is feasible even if the function space is infinite. Since our goal is not to explain the mathematics and mechanism of machine learning, further details on why there is a finite bound on the generalization error are not mentioned here. For readers who are interested in the theory of machine learning, such as Hoeffding's inequality that gives a probability upper bound, Vapnik–Chervonenkis (VC) dimension and VC generalization bound, please refer to the textbooks~\cite{abu2012learning}.

\subsection{Metrics}
Choosing an appropriate numeric evaluation metric for optimization is crucial. Different evaluation metrics are applied to different scenarios and problems.

\subsubsection{Supervised learning}
In classification problems, accuracy, precision/positive predictive value (PPV), recall/sensitivity, specificity, and the F1 score are usually used. We use a confusion matrix to show the relation between these metrics~\ref{tab:metrics}. 

\begin{table*}[htbp]
\centering
\begin{tabular}{cc|c|c|c}
\toprule
 &  & \multicolumn{2}{c|}{Predicted} &  \\ \cline{3-4}
 &  & True & False &  \\ \hline
\multicolumn{1}{c|}{\multirow{2}{*}{Actual}} & True & True positive (TP) & \begin{tabular}[c]{@{}c@{}}False negative (FN)\\ Type II error\end{tabular} & Recall = Sensntivity = $\frac{\mathrm{TP}}{\mathrm{TP+FN}}$ \\ \cline{2-5} 
\multicolumn{1}{c|}{} & False & \begin{tabular}[c]{@{}c@{}}False positive (FP)\\ Type I error\end{tabular} & True negative (TN) & Specificity = $\frac{\mathrm{TN}}{\mathrm{TN+FP}}$ \\ \hline
 &  & Precision = $\frac{\mathrm{TP}}{\mathrm{TP+FP}}$ &  & \begin{tabular}[c]{@{}c@{}}Accuracy = $\frac{\mathrm{TP+TN}}{\mathrm{TP+TN+FP+FN}}$ \\ F1 = $\frac{\mathrm{2} \times \mathrm{Precision} \times \mathrm{Recall}}{\mathrm{Precision+Recall}}$ \end{tabular} \\
\bottomrule
\end{tabular}
\caption{Commonly-used metrics in machine learning.}
\label{tab:metrics}
\end{table*}

\begin{figure}[htbp]
\centering
\includegraphics[height=0.4\linewidth]{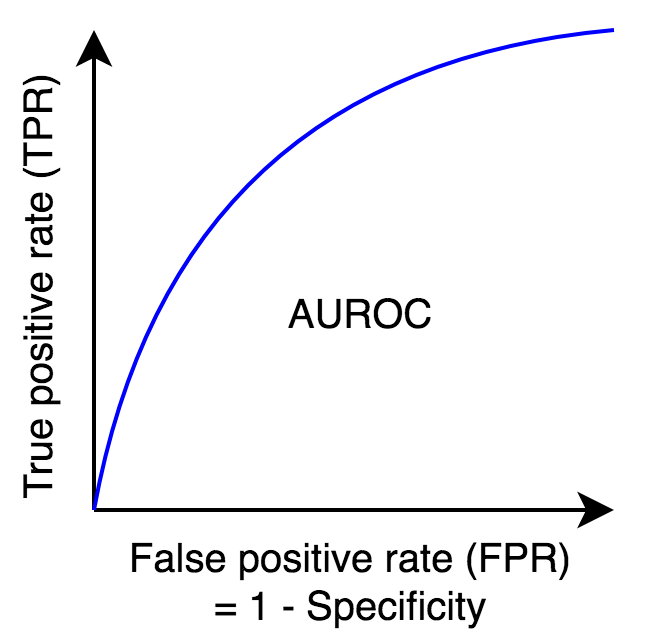}
\caption{Example of AUROC.}
\label{fig:auc}
\end{figure}

The area under receiver operating curve (AUROC) is a very common metric, which sums up the area under the curve in the plot with $x$-axis of false positive rate (FPR, also known as 1-specificity), and $y$-axis of true positive rate (TPR)~\ref{fig:auc}. FPR and TPR values may change based on the threshold of your subjective choice.

In a regression problem, the adjusted R-squared value is commonly used for evaluation. The R-squared value, also known as the coefficient of determination, follows the equation and is defined by the total sum of squares (SStot) and the residual sum of squares (SSres). The detailed equations are as follows: 

$$\mathrm{R}^2 = 1 - \frac{SS_{res}}{SS_{tot}} = 1 - \frac{\sum_{i=1}^m (y_i - f(x_i))^2}{\sum_{i=1}^m (y_i - \hat{y_i})^2}$$
$$\mathrm{Adjusted~R}^2 = 1 - \frac{(1-\mathrm{R}^2)(m-1)}{m-n-1}$$

There are also other metrics for regression, such as Akaike information criterion (AIC) and Bayesian information criterion (BIC), for different study purposes.

\subsubsection{Unsupervised learning}
Since there are no ground truth labels for unsupervised scenarios, evaluation metrics of unsupervised learning settings are relatively difficult to define and usually depend on the algorithms in question. For example, the Calinski-Harabaz index and silhouette coefficient have been used to evaluate k-means clustering. Reconstruction error is used for autoencoder, a kind of neural network architecture for learning data representation.

\subsection{Model Validation}
The next step after deciding the algorithm is to get your data ready for training a model for your task. In practice, we split the whole dataset into three pieces:
\begin{itemize}
    \item Training set for model training. You will run the selected machine learning algorithm only on this subset.
    \item Development (a.k.a. dev, validation) set, also called hold-out, for parameter tuning and feature selection. This subset is only for optimization and model validation.
    \item Testing set for evaluating model performance. We only apply the model for prediction here, but won’t change any content in the model at this moment.
\end{itemize}
There are a few things that we need to keep in mind:
\begin{itemize}
    \item It is better to have your training, dev and testing sets all from the same data distribution instead of having them too different (e.g. training/dev on male patients but testing on female patients), otherwise you may face the problem of overfitting, in which your model will fit the data too well in training or dev sets but find it difficult to generalize to the test data. In this situation, the trained model will not be able to be applied to other cases.
    \item It is important to prevent using any data in the dev set or testing set for model training. Test data leakage, i.e. having part of testing data while training phase, may cause the overfitting of the model to your test data and erroneously gives you a high performance but a bad model.
\end{itemize}
There is no consensus on the relative proportions of the three subsets. However, people usually split out 20-30\% of the whole dataset for their testing set. The proportion can be smaller if you have more data.

\subsubsection{Cross-validation}
The other commonly used approach for model validation is $k$-fold cross validation (CV). The goal of $k$-fold CV is to reduce the overfitting of the initial training set by further training several models with the same algorithm but with different training/dev set splitting.

\begin{figure}[htbp]
\centering
\includegraphics[height=0.4\linewidth]{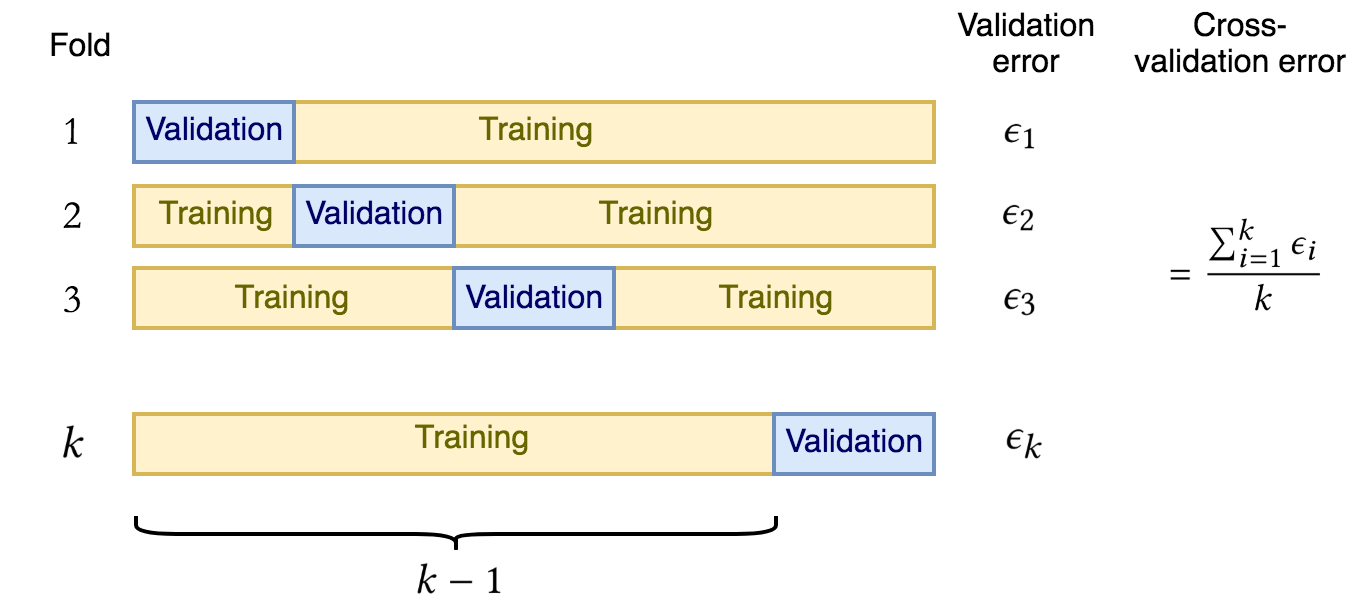}
\caption{K-fold cross-validation.}
\label{fig:cv}
\end{figure}

In $k$-fold CV, we split the whole dataset into $k$ folds and train the model $k$ times. In each training, we iteratively leave one different fold out for validation, and train on the remaining $k-1$ folds. The final error is the average of errors over $k$ times of training~\ref{fig:cv}. In practice, we usually use k=5 or 10. The extreme case for n cases is n-fold CV, which is also called leave-one-out CV (LOOCV).

Please keep in mind that the testing set is completely excluded from the process of CV. Only training and dev sets are involved in this process.

\subsection{Diagnostics}
After the first iteration of model training and evaluation, you may find that the trained model does not perform well on the unseen testing data. To address the issue of error in machine learning, we need to conduct some diagnostics regarding bias and variance in the model in order to achieve a model with low bias and low variance.

\subsubsection{Bias and variance}
The bias of a model is the difference between the expected prediction and the correct model that we try to predict for given data points. That is, it is the algorithm's error rate on training set. This is an underfitting problem, which the model can't capture the trend of the data well due to excessively simple model, and one potential solution is to make the model more complex, which can be done by reducing regularization (section~\ref{s262}), or configuring and adding more input features. For example, stacking more layers if you are using a deep learning approach. However, it is possible that the outcome of complex model is high variance.

The variance of a model is the variability of the model prediction for given data points. It is the model error rate difference between training and dev sets. Problems of high variance are usually related to the issue of overfitting. i.e. hard to generalize to unseen data. The possible solution is to simplify the model, such as using regularization, reducing the number of features, or add more training data. Yet the simpler model may also suffer from the issue of high bias.

High bias and high variance can happen simultaneously with very bad models. To achieve the optimal error rate, a.k.a. Bayes error rate, which is an unavoidable bias from the most optimized model, we need to do iterative experiments to find the optimal bias and variance tradeoff.

\begin{figure}[htbp]
\centering
\includegraphics[width=0.95\linewidth]{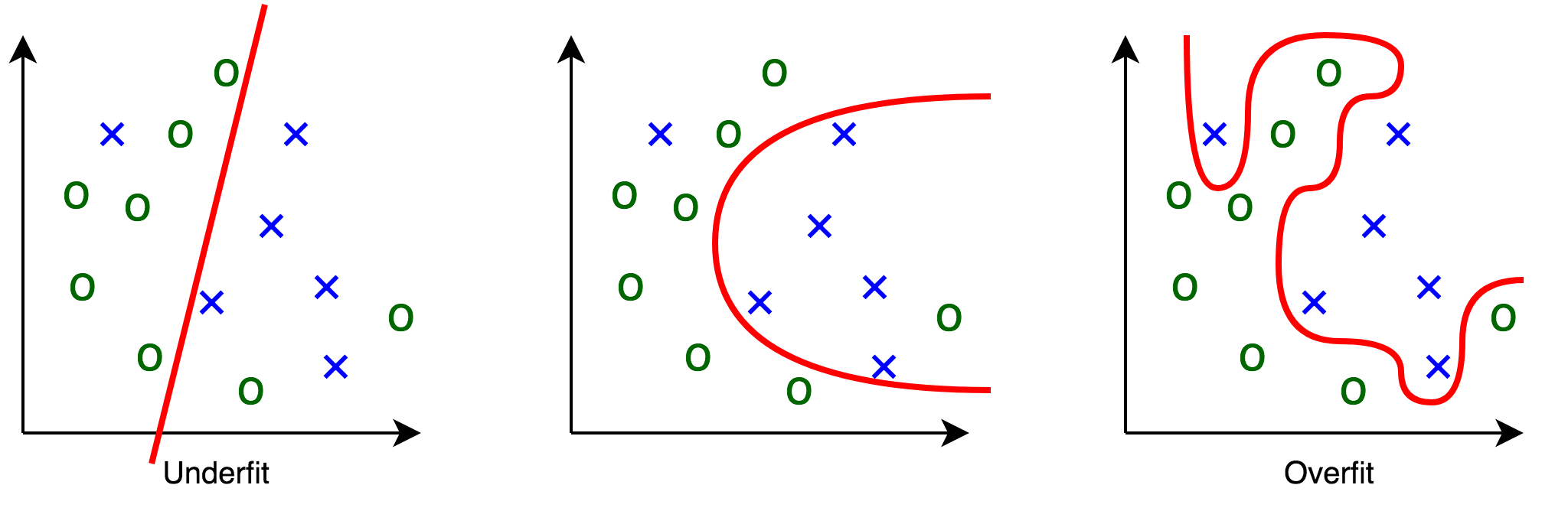}
\includegraphics[width=1.0\linewidth]{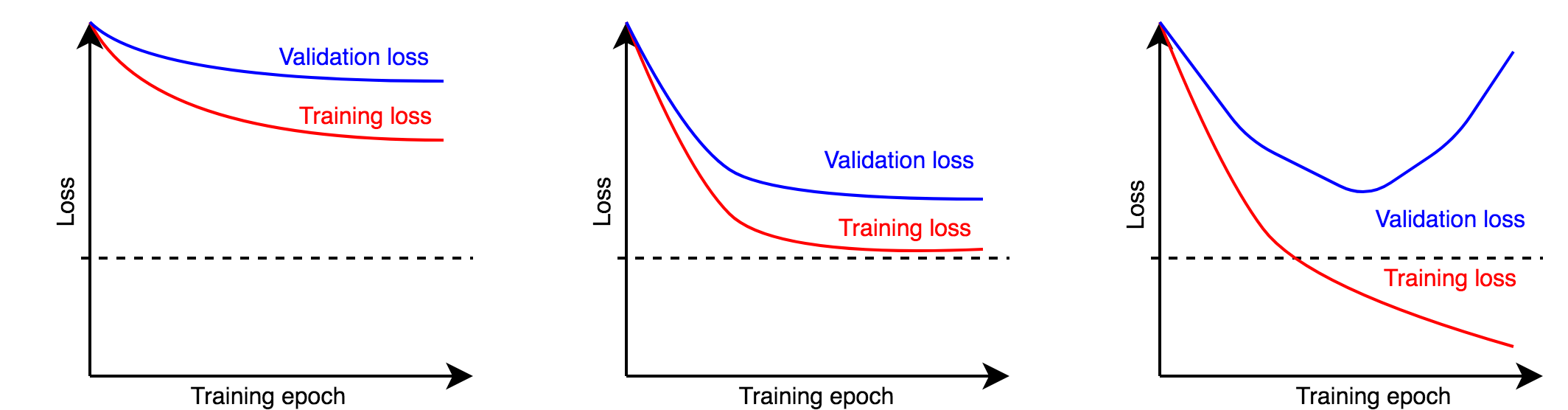}
\caption{Bias and variance.}
\label{fig:dx}
\end{figure}

Finally, a good practice of investigating bias and variance is to plot the informative learning curve with training and validation errors. In Figure~\ref{fig:dx} and Table~\ref{tab:bv} we demonstrate a few cases of diagnostics as examples.

\begin{table*}[htbp]
\centering
\begin{tabular}{cccc}
\toprule
 & Training error & Validation error & Approach \\
\midrule
High bias & High & Low & Increase complexity \\
\midrule
High variance & Low & High & \begin{tabular}[c]{@{}c@{}}Decrease complexity\\ Add more data\end{tabular} \\
\bottomrule
\end{tabular}
\caption{The characteristic of high bias and high variance.}
\label{tab:bv}
\end{table*}

\subsubsection{Regularization}
\label{s262}
The goal of regularization is to prevent model overfitting and high variance. The most common regularization techniques include Least absolute shrinkage and selection operator (LASSO regression, L1-regularization)~\cite{tibshirani1996regression}, ridge regression (L2-regression)~\cite{hoerl1970ridge}, and elastic net regression (a linear combination of L1 and L2 regularization)~\cite{zou2005elasticnet}.

In practice, we add a weighted penalty term $\lambda$ to the cost function as a regularization. For L1-regularization, we add the absolute value of the magnitude of coefficient as penalty term, and in L2-regularization we add the squared value of magnitude instead (Table~\ref{tab:lr}).

\begin{table*}[htbp]
\centering
\begin{tabular}{cc}
\toprule
Regularization & Equation \\
\midrule
L1 (LASSO) & $\sum_{i=1}^m (y_i - \sum_{j=1}^n \beta_j x_{ij})^2 + \lambda \sum_{j=1}^n | \beta_j|$ \\
L2 (Ridge) & $\sum_{i=1}^m (y_i - \sum_{j=1}^n \beta_j x_{ij})^2 + \lambda \sum_{j=1}^n \beta_j^2$ \\
\bottomrule
\end{tabular}
\caption{L1 and L2-regularized logistic regression.}
\label{tab:lr}
\end{table*}

L1-regularization is also a good technique for feature selection since it can "shrink" the coefficients of less important features to zero and remove them. In contrast, L2-regularization just makes the coefficients smaller, but not to zero.

\subsection{Error analysis}
It is an important practice to construct your first prediction pipeline as soon as possible and iteratively improve its performance by error analysis. Error analysis is a critical step to examine the performance between your model and the optimized one. To do the analysis, it is necessary to manually go through some erroneously predicted data from the dev set. 

The error analysis can help you understand potential problems in the current algorithm setting. For example, the misclassified cases usually come from specific classes (e.g. patients with cardiovascular issues might get confused with those with renal problems since there are some shared pathological features between two organ systems) or inputs with specific conditions~\cite{weng2017medical}. Such misclassification can be prevented by changing to more complex model architecture (e.g. neural networks), or adding more features (e.g. combining word- and concept-level features), in order to help distinguish the classes.

\subsection{Ablation analysis}
Ablation analysis is a critical step for identifying important factors in the model.
Once you obtain an ideal model, it is necessary to compare it with some simple but robust models, such as linear or logistic regression model. This step is also essential for research projects, since the readers of your work will want to know what factors and methods are related to the improvement of model performance. For example, the deep learning approach of clinical document deidentification outperforms traditional natural language processing approach. In the paper of using neural network for deidentification~\cite{dernoncourt2017identification}, the authors demonstrate that the character-level token embedding technique had the greatest effect on model performance, and this became the critical factor of their study.

\section{Learning Algorithms}
\label{s3}
In this section, we briefly introduce the concepts of some algorithm families that can be used in the clinical prediction tasks. For supervised learning, we will discuss linear models, tree-based models and SVM. For unsupervised learning, we will discuss the concepts of clustering and dimensionality reduction algorithms. We will skip the neural network method in this chapter. Please refer to programming tutorial part 3 or deep learning textbook for further information~\cite{goodfellow2016deep}.

\subsection{Supervised learning}
\subsubsection{Linear models}
Linear models are commonly used not only in machine learning but also in statistical analysis. They are widely adopted in the clinical world and can usually be provided as baseline models for clinical machine learning tasks. In this class of algorithms, we usually use linear regression for regression problems and logistic regression for classification problems.

The pros of linear models include their interpretability, less computation, as well as less complexity comparing to other classical machine learning algorithms. The cons of them are their inferior performance. However, these are common trade-off features in model selection. It is still worthwhile to start from this simple but powerful family of algorithms.

\subsubsection{Tree-based models}
Tree-based models can be used for both regression and classification problems.
Decision tree, also known as classification and regression trees (CART), is one of the most common tree-based models~\cite{breiman2017classification}. It follows the steps below to find the best tree:
\begin{itemize}
    \item It looks across all possible thresholds across all possible features and picks the single feature split that best separates the data
    \item The data is split on that feature at a specific threshold that yields the highest performance
    \item It iteratively repeats the above two steps until reaching the maximal tree depth, or until all the leaves are pure
\end{itemize}
There are many parameters that should be considered while using the decision tree algorithm. The following are some important parameters:
\begin{itemize}
    \item Splitting criteria: by Gini index or entropy
    \item Tree size: tree depth, tree pruning
    \item Number of samples: minimal samples in a leaf, or minimal sample to split a node
\end{itemize}
The biggest advantage of a decision tree is providing model interpretability and actionable decision. Since the tree is represented in a binary way, the trained tree model can be easily converted into a set of rules. For example, in the paper the authors utilized CART to create a series of clinical rules~\cite{fonarow2005risk}. However, decision trees may have the issue of high variance and yield an inferior performance.

Random forest is another tree-based algorithm that combines the idea of bagging and subsampling features~\cite{breiman2001random}. In brief, it tries to ensemble the results and performances of a number of decision trees that were built by randomly selected sets of features. The algorithm can be explained as follows:
\begin{itemize}
    \item Pick a random subset of features
    \item Create a bootstrap sample of data (randomly resample the data)
    \item Build a decision tree on this data
    \item Iteratively perform the above steps until termination
\end{itemize}
Random forest is a robust classifier that usually works well on most of the supervised learning problems, but a main concern is model interpretability.
There are also other tree-based models such as adaptive boosting (Adaboost) and gradient boosting algorithms, which attempt to combine multiple weaker learners into a stronger model~\cite{freund1999short,friedman2001greedy}.

\subsubsection{Support vector machine (SVM)}
\begin{figure}[htbp]
\centering
\includegraphics[height=0.4\linewidth]{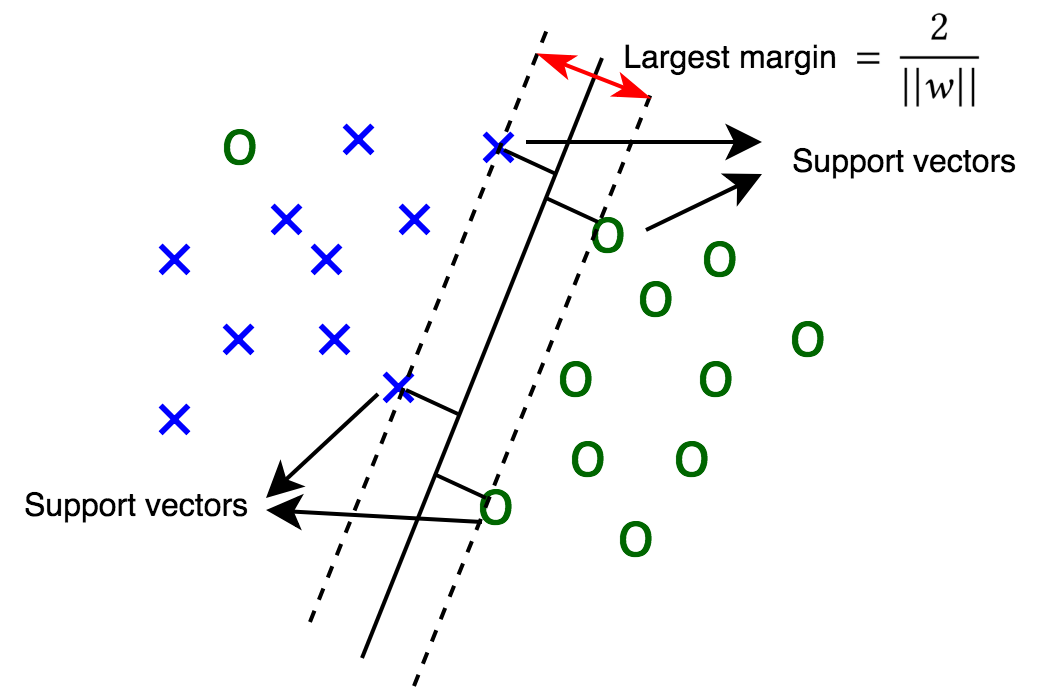}
\caption{Hyperplane of SVM to linearly separate samples.}
\label{fig:svm}
\end{figure}

SVM is a very powerful family of machine learning algorithms~\cite{cortes1995support}. The goal of SVM is trying to find a hyperplane (e.g. a line in 2D, a plane in 3D, or a $n$-dimension structure in a $n+1$ dimensions space) to separate data points into two sides, and the hyperplane has to maximize the minimal distance from the sentinel data points, support vectors, to the hyperplane~\ref{fig:svm}.

\begin{figure}[htbp]
\centering
\includegraphics[width=1.0\linewidth]{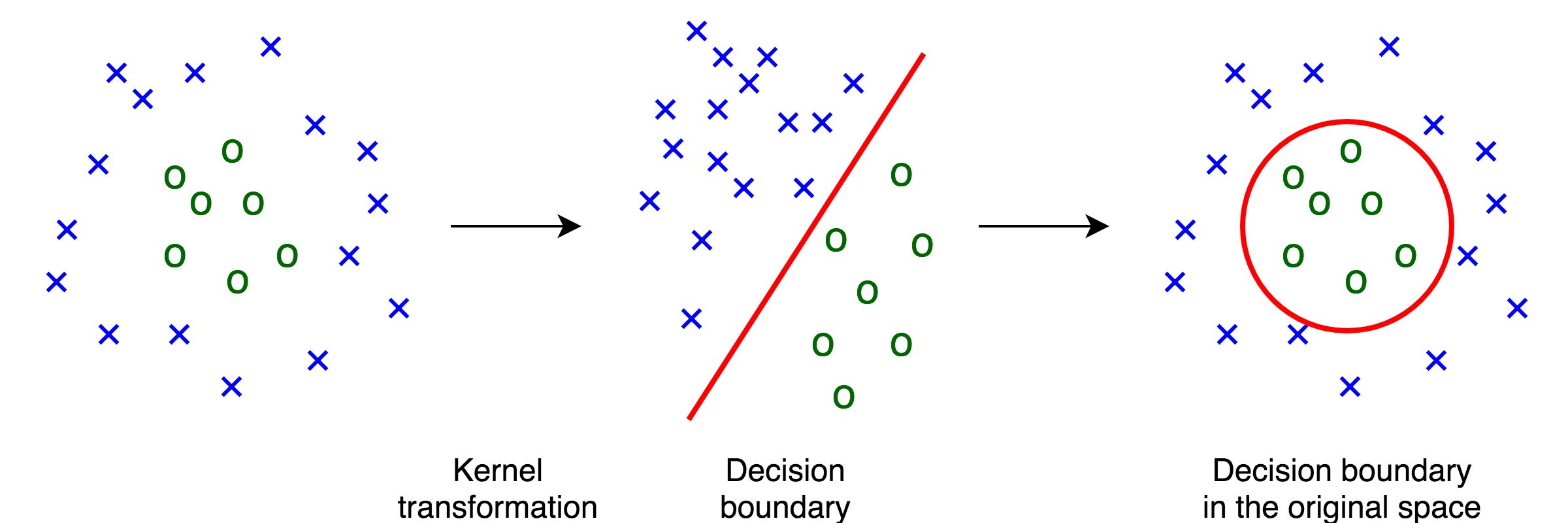}
\caption{Kernel trick of SVM.}
\label{fig:kernel}
\end{figure}

SVM also works for non-linear separable data. It uses a technique called ``kernel trick'' that linearly splits the data in another vector space, then converts the space back to the original one later~\ref{fig:kernel}. The commonly used kernels include linear kernel, radial basis function (RBF) kernel and polynomial kernel.

Regarding the optimization, we used hinge loss to train SVM.
The pros of using SVM is its superior performance, yet the model's inferior interpretability limits its applications in the healthcare domain.

\subsection{Unsupervised learning}
In the previous section, we mentioned that the goal of unsupervised learning is to discover hidden patterns inside data. We can use clustering algorithms to aggregate data points into several clusters and investigate the characteristics of each cluster. We can also use dimensionality reduction algorithms to transform a high-dimensional into a smaller-dimensional vector space for further machine learning steps.

\subsubsection{Clustering}
$K$-means clustering, Expectation-Maximization (EM) algorithm, hierarchical clustering are all common clustering methods. In this section, we will just introduce k-means clustering. The goal of $k$-means clustering is to find latent groups in the data, with the number of groups represented by the variable $k$.

\begin{figure}[htbp]
\centering
\includegraphics[width=1.0\linewidth]{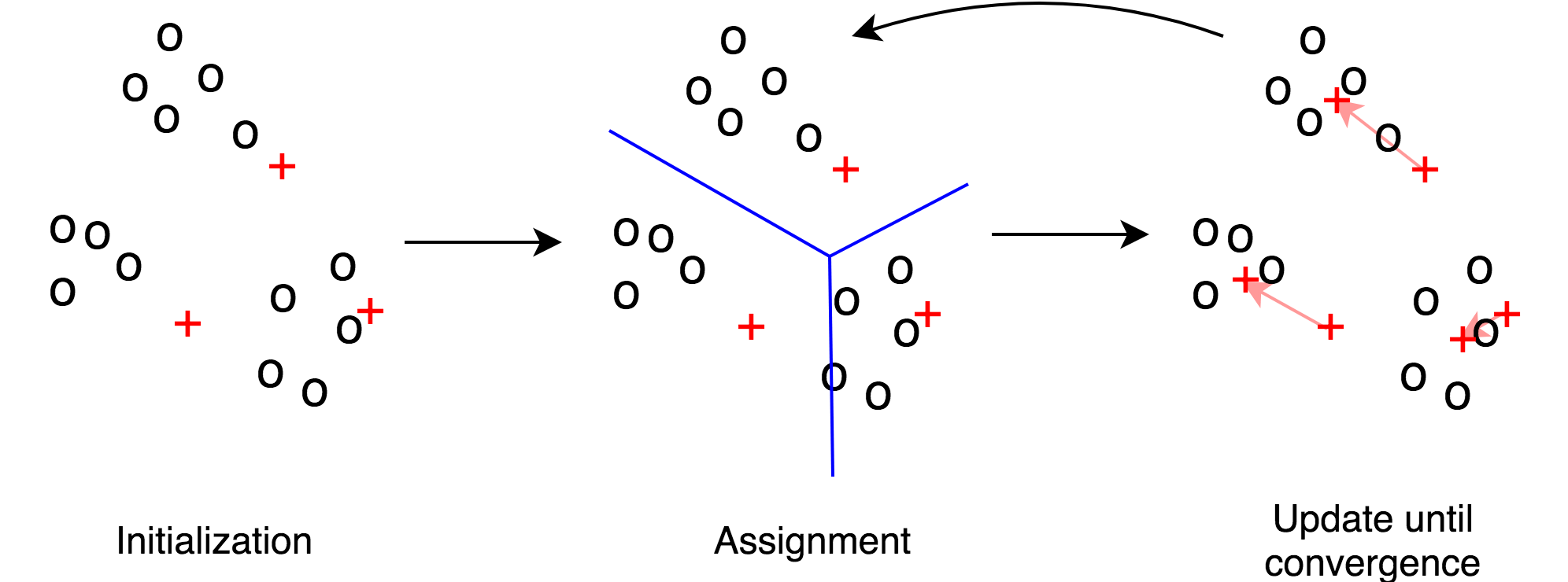}
\caption{Steps of $k$-means clustering.}
\label{fig:km}
\end{figure}

The simplified steps of $k$-means clustering are (Figure~\ref{fig:km}):
\begin{itemize}
    \item Randomly initializing $k$ points as the centroids of the $k$ clusters
    \item Assigning data points to the nearest centroid and forming clusters
    \item Recomputing and updating centroids based on the mean value of data points in the cluster
    \item Repeating step 2 and 3 until convergence
\end{itemize}

The $k$-means algorithm is guaranteed to converge to a final result. However, this converged state may be local optimum and therefore need to experiment several times to explore the variability of results.

The obtained final $k$ centroids, as well as the cluster labels of data points can all serve as new features for further machine learning tasks, as well be shown in Section 9 of the ``Applied Statistical Learning in Python'' chapter.
Regarding choosing the cluster number $k$, there are several techniques for $k$ value validation. The most common methods include elbow method, silhouette coefficient, also Calinski-Harabaz index. However, it is very useful to decide $k$ if you already have some clinical domain insights about potential cluster number.

\subsubsection{Dimensionality reduction}
While dealing with clinical data, it is possible that you need to face with a very high-dimensional but sparse dataset. Such characteristics may decrease the model performance even if you use top performing machine algorithms such as SVM, random forest or even deep learning due to the risk of overfitting. A potential solution is to utilize the power of dimensionality reduction algorithms to convert the dataset into lower dimensional vector space.Principal component analysis (PCA) is a method that finds the principal components of the data by transforming data points into a new coordinate system~\cite{jolliffe2011principal}. The first axis of the new coordinate system corresponds to the first principal component (PC1), which explains the most variance in the data and can serve as the most important feature of the dataset.

PCA is a linear algorithm and therefore it is hard to interpret the complex polynomial relationship between features. Also, PCA may not be able to represent similar data points of high-dimensional data that are close together since the linear algorithm does not consider non-linear manifolds.

The non-linear dimensionality reduction algorithm, t-Distributed Stochastic Neighbor Embedding (t-SNE), becomes an alternative when we want to explore or visualize the high-dimensional data~\cite{maaten2008visualizing}. t-SNE considers probability distributions with random walk on neighborhood graphs on the curved manifold to find the patterns of data. Autoencoder is another dimensionality reduction algorithm based on a neural network architecture that aims for learning data representation by minimizing the difference between the input and output of the network~\cite{rumelhart1988learning,hinton2006reducing}.

The dimensionality reduction algorithms are good at representing multi-dimensional data. Also, a smaller set of features learned from dimensionality reduction algorithms may not only reduce the complexity of the model, but also decrease model training time, as well as inference (classification/prediction) time.

\section{Pitfalls and Limitations}
\label{s4}
Machine learning is a powerful technique for healthcare research. From a technical and algorithmic perspective, there are many directions that we can undertake to improve methodology, such as generalizability, less supervision, multimodal and multitask training~\cite{weng2019multimodal}, or learning temporality and irregularity~\cite{xiao2018opportunities}. 

However, there are some pitfalls and limitations about utilizing machine learning in healthcare that should be considered while model development~\cite{chen2019develop}. For example, model biases and fairness is a critical issue since the training data we use are usually noisy and biased~\cite{caruana2015intelligible,ghassemi2018opportunities}. We still need human expert to validate, interpret and adjust the models. Model interpretability is also an important topic from the aspects of (1) human-machine collaboration and (2) building a human-like intelligent machine for medicine~\cite{girkar2018predicting}. Causality is usually not being addressed in most of the clinical machine learning research, yet it is a key of clinical decision making. We may need more complicated causal inference algorithms to answer clinical causal questions.

We also need to think more about how to deploy the developed machine learning models into clinical workflow. How to utilize them to improve workflow~\cite{horng2017creating,chen2019augmented}, as well as integrate all information acquired by human and machine, to transform them into clinical actions and improve health outcomes are the most important things that we should consider for future clinician-machine collaboration.

\section{Programming Exercise}
\label{s5}
We provide three tutorials for readers to have some hands-on exercises of learning basic machine learning concepts, algorithms and toolkits for clinical prediction tasks. They can be accessed through Google colab and Python Jupyter notebook with two real-world datasets:
\begin{itemize}
    \item Breast Cancer Wisconsin (Diagnostic) Database
    \item Preprocessed ICU data from PhysioNet Challenge 2012 Database
\end{itemize}
The learning objectives of these tutorial include:
\begin{itemize}
    \item Learn how to use Google colab / Jupyter notebook
    \item Learn how to build and diagnose machine learning models for clinical classification and clustering tasks
\end{itemize}
In part 1, we will go through the basic of machine learning concepts through classification problems. In part 2, we will go deeper into unsupervised learning methods for clustering and visualization. In part 3, we will discuss more about deep neural networks. Please check the link of tutorials in the Appendix.

\section{Conclusion}
\label{s6}
In summary, machine learning is an important and powerful technique for healthcare research. In this chapter, we have shown readers how to reframe a clinical problem into appropriate machine learning tasks, select and adjust an algorithm for model training, perform model diagnostics and error analysis, as well as model results and interpretation. The concepts and tools described in this chapter aim  to allow the researcher to better understand how to conduct a machine learning project for clinical predictive analytics.

\section*{Programming Tutorial Appendix}
The tutorials mentioned in this chapter  available in the GitHub repository: \url{https://github.com/ckbjimmy/2018_mlw.git}.

\bibliography{references}
\bibliographystyle{icml2019}

\end{document}